\documentclass[letterpaper, 10 pt, conference]{ieeeconf}  

\IEEEoverridecommandlockouts                              

\usepackage{booktabs}
\usepackage{graphicx}
\usepackage{amsmath} 
\usepackage{amssymb}
\usepackage{textcomp}
\usepackage{booktabs}
\usepackage{marvosym}
\usepackage{threeparttable}
\usepackage{xcolor} 
\usepackage[hypcap=true]{caption}
\usepackage{cite}
\usepackage{hyperref} 
\usepackage{multirow}

\title{\LARGE \bf
Reusing Attention for One-stage Lane Topology Understanding
}

\author{Yang Li$^{*1, 4}$, Zongzheng Zhang$^{*1, 2}$, Xuchong Qiu$^{*2}$, Xinrun Li$^{2}$, Ziming Liu$^{2}$,  Leichen Wang$^{2}$, \\Ruikai Li$^{5}$, Zhenxin Zhu$^{1}$, Huan-ang Gao$^{1}$, Xiaojian Lin$^{1}$, Zhiyong Cui$^{5}$, Hang Zhao$^{3}$, and Hao Zhao$^{1}\textsuperscript{\Letter}$
\thanks{$^{1}$Institute for AI Industry Research, Tsinghua University.}%
\thanks{$^{2}$Bosch Corporate Research, China.}%
\thanks{$^{3}$Institute for Interdisciplinary Information Sciences, Tsinghua University.}
\thanks{$^{4}$Department of Computer Science, ETH Zürich.}%
\thanks{$^{5}$State Key Lab of Intelligent Transportation System, Beihang University.}%
\thanks{$*$ Equal contribution.}
\thanks{\textsuperscript{\Letter} Corresponding to zhaohao@air.tsinghua.edu.cn}
}

\begin{document}

\makeatletter
\let\@oldmaketitle\@maketitle
\renewcommand{\@maketitle}{\@oldmaketitle%
    \centering
    \begingroup
    \setcounter{figure}{0}
    \captionsetup{type=figure}
    \includegraphics[width=0.9\linewidth]{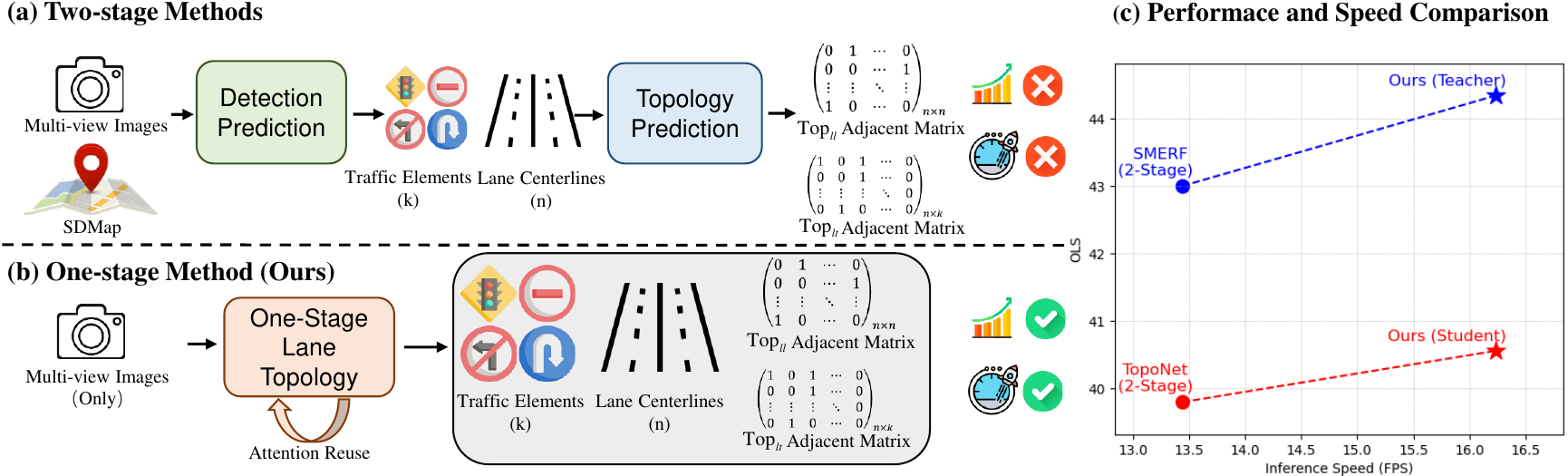}
    \captionof{figure}{(a) Two-stage methods typically use multi-view images and auxiliary SD maps as input for the detection-prediction module to detect traffic elements and lane centerlines. Topology prediction is then performed to generate relationship matrix. (b) Our one-stage method simultaneously performs detection and topology prediction through attention reuse. (c)
    Our method achieves \textbf{higher performance} and \textbf{faster inference speed} both \textcolor{blue}{with SD maps} and \textcolor{red}{without SD maps}.}
    \vspace{-10pt}
    \label{fig:teaser}
    \endgroup

}
\makeatother

\maketitle
\thispagestyle{empty}
\pagestyle{empty}


\begin{abstract}
Understanding lane toplogy relationships accurately is critical for safe autonomous driving. However, existing two-stage methods suffer from inefficiencies due to error propagations and increased computational overheads. To address these challenges, we  propose a one-stage architecture that simultaneously predicts traffic elements, lane centerlines and topology relationship, improving both the accuracy and inference speed of lane topology understanding for autonomous driving. Our key innovation lies in reusing intermediate attention resources within distinct transformer decoders. This approach effectively leverages the inherent relational knowledge within the element detection module to enable the modeling of topology relationships among traffic elements and lanes without requiring additional computationally expensive graph networks. Furthermore, we are the first to demonstrate that knowledge can be distilled from models that utilize standard definition (SD) maps to those operates without using SD maps, enabling superior performance even in the absence of SD maps. Extensive experiments on the OpenLane-V2 dataset show that our approach outperforms baseline methods in both accuracy and efficiency, achieving superior results in lane detection, traffic element identification, and topology reasoning. Our code is available at \href{https://github.com/Yang-Li-2000/one-stage.git}{https://github.com/Yang-Li-2000/one-stage.git}.

\end{abstract}

\section{INTRODUCTION}
Scene understanding plays a pivotal role in autonomous driving. To ensure safe navigation, particularly in challenging scenarios, it is essential to achieve both accurate detection and topology reasoning simultaneously, as these capabilities are critical for effective planning and control of autonomous vehicles \cite{openlanev2,tian2023unsupervised}.

However, existing road topology reasoning methods~\cite{smerf, STSU, vectormapnet, maptr, toponet}, which are two-stage, often struggle to achieve both tasks effectively due to the inherent trade-off between detection and topology reasoning. In these approaches, the network first detects traffic elements and lane centerlines and then reasons about their topological relationships (Fig.~\ref{fig:teaser}(a)). This sequential process can lead to error propagation, where inaccuracies in the detection stage adversely affect the topology reasoning stage. Furthermore, optimizing for one task may result in features that are less suitable for the other task, creating a bottleneck in achieving robust scene understanding. 

To address the limitations of two-stage framework, we propose a novel one-stage architecture that simultaneously performs road elements detection and topology reasoning (Fig.~\ref{fig:teaser}(b)). Our approach enables more efficient optimization of both tasks, fostering better feature sharing and reducing the risk of error propagation, ultimately leading to more accurate scene understanding. 

In addition to achieving higher accuracy, our one-stage architecture significantly improves inference speed, achieving a 17\% reduction in inference time (Fig.~\ref{fig:teaser}(c)). Traditional two-stage methods, such as TopoNet \cite{toponet}, first detect centerlines and traffic elements before constructing a graph neural network (GNN) to reason about topology. This process involves time-consuming steps, including GNN construction and prediction, as well as a large number of parameters, which increase computational overhead. In contrast, our proposed method eliminates these inefficiencies by leveraging lightweight layers to perform topology reasoning directly, resulting in faster inference.

The key innovation of our approach lies in reusing attention resources directly from transformer decoder for topology understanding. Specifically, we take out queries and keys from intermediate layers of transformer decoder, along with the last-layer decoder outputs, performs projection, pair-wise concatenation, and gated sum similar to EGTR \cite{egtr}. 

However, our approach significantly differs from EGTR in both methodology and application. While EGTR focuses on extracting relationships from a single transformer decoder by leveraging self-attention weights within a unified object detection framework, we propose a novel cross-decoder topology reasoning mechanism. Specifically, we take features from two distinct transformers — one dedicated to traffic elements (using perspective-view features from the front camera) and the other to lane centerlines (using Bird's-Eye-View Features constructed from multi-view inputs). This dual-decoder architecture allows us to reason about topology relationships across two fundamentally different feature spaces, which is a significant departure from EGTR's single-decoder design. 

To the best of our knowledge, we are the first to demonstrate that such cross-decoder topology reasoning is feasible, particularly in the context of autonomous driving, where BEV and perspective-view features are inherently complementary but challenging to unify. This approach not only extends the applicability of transformer-based topology reasoning but also addresses the unique challenges of lane and traffic element understanding in complex driving scenes.


Additionally, we propose distilling knowledge from SD-map-based models into SD-map-free models to enhance accuracy when SD maps are unavailable. While SD maps offer higher availability and lower costs compared to High Definition (HD) maps \cite{smerf}, they are not always accessible or up-to-date, particularly in remote areas, regions with frequent road or infrastructure changes, newly developed areas, as well as underground tunnels and parking garages.


In summary, our key contributions are as follows:
\begin{itemize}
    \item A novel one-stage architecture for simultaneous detection and topology reasoning, which mitigates error propagation and optimizes feature sharing to enhance scene understanding.
    \item A cross-decoder topology reasoning mechanism, which leverages separate transformer decoders for traffic elements and lane centerlines, enabling effective reasoning across distinct feature spaces.
    \item A knowledge distillation framework for SD-map-free models, which transfers knowledge from SD-map-based models to improve accuracy in scenarios where SD maps are unavailable.
    \item Comprehensive experiments demonstrating superior performance, showing that our approach achieves higher accuracy and faster inference both with and without SD maps. 
\end{itemize}

\begin{figure*}[t!]
  \centering
   \includegraphics[width=0.8\linewidth]{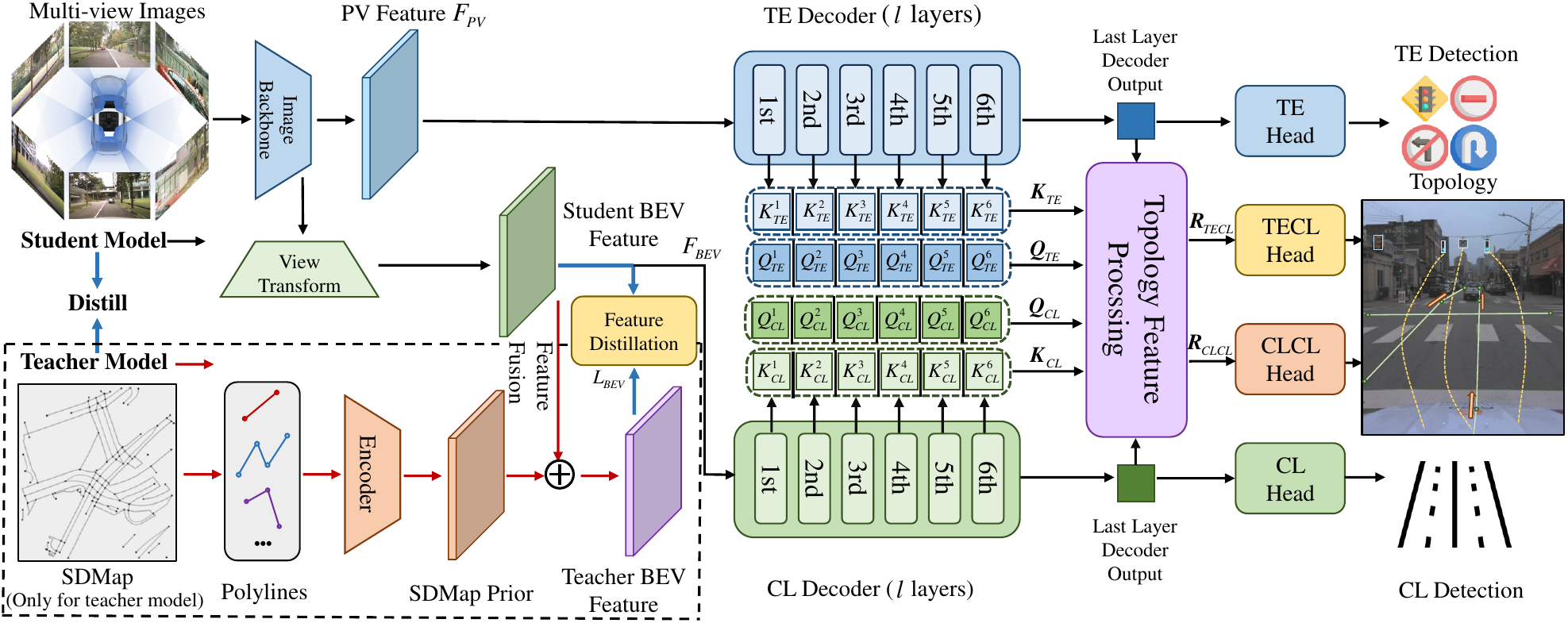}
   \caption{\textbf{Overall Architecture.} We propose a novel one-stage method where intermediate queries ($\textbf{Q}_{TE}$ and $\textbf{Q}_{CL}$) and keys ($\textbf{K}_{TE}$ and $\textbf{K}_{CL}$) are extracted from each self-attention layer within the traffic elements (TE) and lane centerlines (CL) decoders, and are subsequently used for topology reasoning.  Concurrently, traffic elements and centerlines are predicted using the final-layer decoder outputs. We further introduce a teacher-student knowledge distillation framework that applies distillation to BEV features ($F_{BEV}$). We abbreviate ``TECL" and ``CLCL" to denote topological relationships between traffic elements and lane centerlines, and between lane centerlines themselves, respectively. 
}
   \label{architecture}
   \vspace{-15pt}
\end{figure*}

\section{Related Work}
\subsection{Lane Topology Construction}
Given sensor data, lane topology construction aims to detect lanes and traffic elements and reason about their relations. Early works focus on road topology generation from bird-eye-views such as aerial images ~\cite{yao2024aerial,zhou2021aerial,zurn2021aerial,jin2024tod3cap}. For onboard sensors, STSU~\cite{STSU} proposes to first detect both static road elements and dynamic objects with a Transformer-based model, then estimate the relations between these detected instances. TopoRoad~\cite{toporoad} better maintains the order of relations between vertices by introducing additional cycle queries. Similarly, Can et al.~\cite{can2023improving} propose to consider the centerlines as cluster centers in object assignment to offer an additional supervision for enhancing relation prediction in road topology. Contrast to previous end-to-end methods, LaneGAP~\cite{lanegap} recovers the topology from a set of lanes by introducing a heuristic-based algorithm. CenterlineDet~\cite{xu2023centerlinedet} and TopoNet~\cite{toponet} propose the respective neural graph models to estimate the centerline topology. TopoMLP~\cite{wu2023topomlp} employs a novel positional embedding to enhance the road topology reasoning. Chameleon~\cite{zhang2025chameleon} combines neuro-symbolic reasoning with VLMs to extract lane topology in a few-shot manner, balancing accuracy and efficiency.

Despite these advances, existing methods all adopt a two-stage framework, where the detector detects vertices and topology predictor estimates relationships separately, resulting in ineffectiveness and inefficiency\cite{liu2023delving}. In contrast to them, we propose a novel one-stage architecture unifying instance detection and relation prediction for lane topology reasoning. 

\subsection{Scene Graph Generation}

Scene Graph Generation (SGG) aims to construct structured representations from visual scenes, where nodes represent objects and edges denote inter-object relationships~\cite{xu2017scene,yang2018graph,tang2020unbiased,chen2022pq,li2023understanding,li2022distance,chen2022cerberus,gao2023semi,li2022toist}. SGG has received increasing attention in computer vision community, due to the huge potential for down-stream visual reasoning tasks. The existing SGG methods can be divided into two groups : two-stage methods and one-stage methods. Two-stage methods~\cite{lu2016visual,xu2017scene,zellers2018neural,woo2018linknet,tang2019learning,koner2020relation,lin2020gps,lu2021context,dhingra2021bgt,li2021bipartite,min2023environment,sudhakaran2023vision,li2023compositional} usually first detect objects from conventional object detectors such as FasterRCNN~\cite{ren2015faster} and YOLO~\cite{redmon2016you}, and then feed all detected objects into the relation prediction model to estimate the relation between each object pair. The separate object detector and relation predictor are trained in a sequential manner. Despite the effectiveness of this paradigm, the inherent nature of separate modules leads to an evident increase of computational complexity in training and testing. 

One-stage methods~\cite{newell2017pixels,liu2021fully,teng2022structured,shit2022relationformer,khandelwal2022iterative,cong2023reltr,li2023understanding,li2022distance},train object detection and relation prediction jointly in an end-to-end manner. Earlier works adopt fully convolutional networks~\cite{lin2020gps}, while recent advances are inspired by query-based DETR~\cite{detr}. More recent methods~\cite{li2022sgtr, cong2023reltr,liao2024uniq,desai2022single,li2023understanding,li2022distance} introduce object queries or triplet queries in SGG modeling for better efficiency. RelTR~\cite{cong2023reltr} introduces paired subject queries and object queries while SGTR~\cite{li2022sgtr} proposes using compositional queries decoupled into subjects, objects, and predicates. EGTR~\cite{egtr}  leverages self-attention in DETR decoders to extract relation graphs. However, these methods operate uniquely on camera perspective views, and cannot handle cross-view road topology reasoning for autonomous driving.

In this work, we build upon the success of one-stage SGG models for relation prediction, and propose a novel one-stage SGG method specifically designed for road ontology estimation, enabling cross-view relation estimation between the front view (PV) and Bird's-Eye-View (BEV) perspectives.

\section{Method}
This section outlines the key components of our architecture, as shown in Fig.~\ref{architecture}. Specifically, We first describe the feature extraction in Sec.~\ref{method_feature_extraction}. Then we present our approach to topology reasoning within a unified one-stage framework in Sec.~\ref{method_topology_reasoning}. Finally, We present our method for knowledge distillation from map-based teachers to map-free students in Sec.~\ref{method_distill}. 

\subsection{Feature Extraction}
\label{method_feature_extraction}

As illustrated in Fig. \ref{architecture}, We use multi-view images as input, from which front-view image features $F_{PV}$ are extracted through an image backbone. To facilitate comparison with the TopoNet~\cite{toponet} baseline in experiments, we adopt the same backbone architecture, consisting of a pre-trained ResNet-50 and a Feature Pyramid Network (FPN)~\cite{lin2017feature}. The extracted image features are then transformed into Bird's-Eye-View (BEV) features $F_{BEV}$ using the same view transform module as in \cite{toponet}. 

\begin{figure*}[t]
  \centering
   \includegraphics[width=0.75\linewidth]{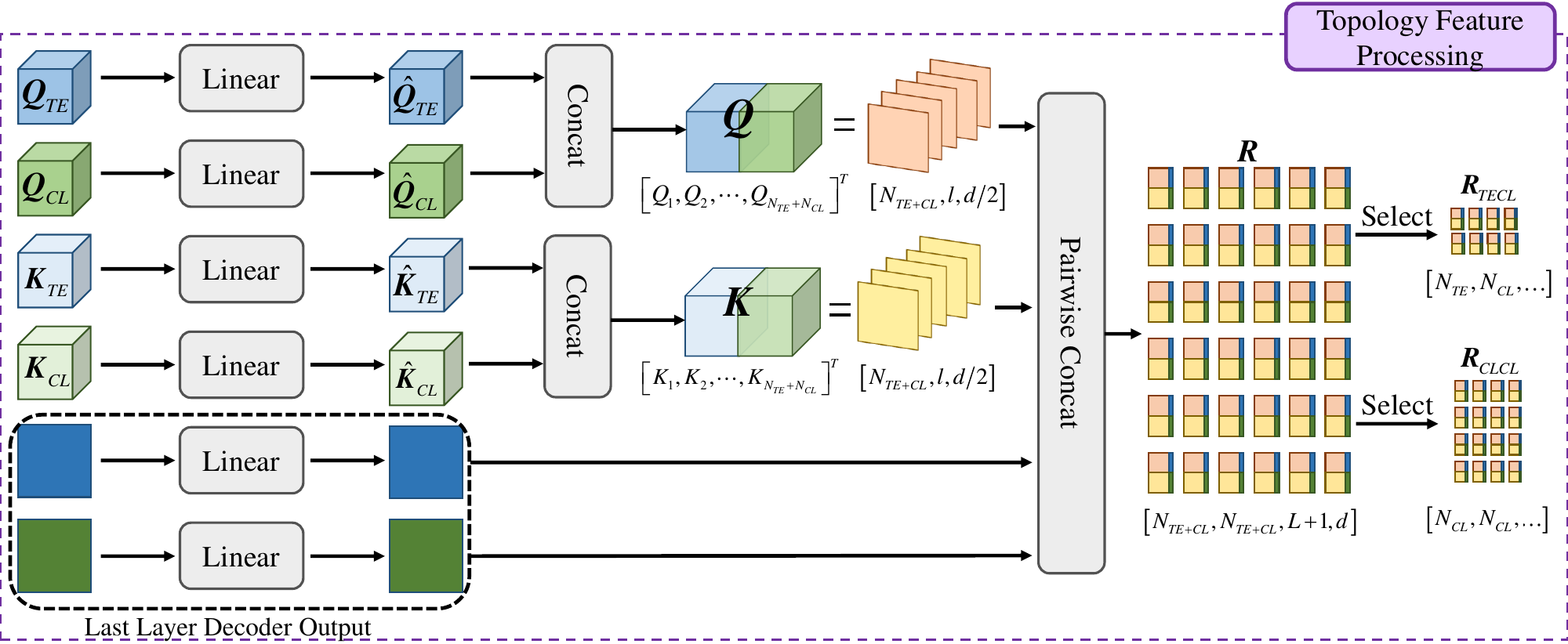}
   \caption{\textbf{Topology Feature Processing.} Intermediate queries and keys extracted from each self-attention layer within the traffic elements (TE) and centerlines (CL) decoders are projected using linear layers and concatenated. Outputs from the last layers of the two decoders are also projected using linear layers. After that, projected features are pairwise concatenated. Finally, task-relevant projected features are selected and fed into the TECL Head and CLCL Head.}
   \label{topology_feature_processing}
   \vspace{-15pt}
\end{figure*}

\subsection{One-Stage Prediction for Topology Reasoning}
\label{method_topology_reasoning}
Our key innovations lies in the proposed one-stage prediction approach, where we are the first to generate road topology predictions simultaneously with detection predictions. In contrast, previous two-stage methods, such as TopoNet~\cite{toponet} and SMERF~\cite{smerf}, first produce detection outputs to construct a graph before inferring topology using GNNs.

From the traffic elements (TE) transformer decoder and lane centerline (CL) transformer decoder, we extract queries and keys from their self-attention layers. By using those extracted queries and keys as input, our model outputs topology predictions. Specifically, the decoders have $L$ layers. As shown in Fig.~\ref{architecture}, the traffic element decoder takes camera front-view features $F_{PV}$ and predicts bounding boxes for traffic elements. The output from the last layer decoder is then fed into the TE Head, which utilizes DETR~\cite{detr} heads for detection. From the self-attention in its each layer $l$, we take out queries $\mathbf{Q}_{TE}^{l}$ and keys $\mathbf{K}_{TE}^{l}$~$\in~\mathbb{R}^{N_{TE} \times 1 \times d}$ where $N_{TE}$ is the number of queries in the traffic element decoder and $d$ is the dimension of transformer layers. Similarly, the BEV feature $F_{BEV}$ is passed throgh the centerline transformer, and the last layer decoder output is input to the CL Head for centerline detection. Simultaneously, from layers in the centerline decoder, we take out queries $\mathbf{Q}_{CL}^{l}$ and keys $\mathbf{K}_{CL}^{l}$~$\in~\mathbb{R} ^{N_{CL} \times 1 \times d}$ as shown in Fig.~\ref{architecture}. These keys and queries are stacked to from $\mathbf{Q}_{TE} $ and $\mathbf{K}_{TE}$  $\in \mathbb{R}^{N_{TE}\times L\times d}$, and $\mathbf{Q}_{CL} $ and $\mathbf{K}_{CL}$~$\in~\mathbb{R}^{N_{CL} \times L \times d}$.

Before performing topology reasoning, we first fuse the extracted queries and keys across camera perspective view and Bird's-Eye-View to get proper features. This process is also illustrated in Fig.~\ref{topology_feature_processing}. When predicting the topology, we first concatenate linearly projected and stacked queries $\mathbf{Q}_{TE}$ and $\mathbf{Q}_{CL}$ to get $\mathbf{Q}$ $\in \mathbb{R}^{(N_{TE} + N_{CL}) \times L \times (d/2)}$. Similarly, we concatenate linearly projected stacked $\mathbf{K}_{TE}$ and $\mathbf{K}_{CL}$ to get $\mathbf{K}$~$\in~\mathbb{R}^{(N_{TE} + N_{CL}) \times L \times (d/2)}$: \vspace{-0.5cm}

\begin{align}
\mathbf{\hat{Q}}_{*}^{l} &= \mathbf{W}_{Q,*}^{l} \mathbf{Q}_{*}^{l} + \mathbf{b}_{Q,*}^{l} 
\end{align}
\begin{align}
\mathbf{\hat{K}}_{*}^{l} &= \mathbf{W}_{K,*}^{l} \mathbf{K}_{*}^{l} + \mathbf{b}_{K,*}^{l} 
\end{align}

\begin{align}
\mathbf{Q} &= \begin{bmatrix}
\mathbf{Q}^{1} \\
\mathbf{Q}^{2} \\
\vdots \\
\mathbf{Q}^{N} \\
\end{bmatrix} = \begin{bmatrix}
\mathbf{\hat{Q}}_{TE}^1 & \mathbf{\hat{Q}}_{TE}^2 & \dots & \mathbf{\hat{Q}}_{TE}^L \\
\mathbf{\hat{Q}}_{CL}^1 & \mathbf{\hat{Q}}_{CL}^2 & \dots & \mathbf{\hat{Q}}_{CL}^L
\end{bmatrix} 
\end{align}

\begin{align}
\mathbf{K} &= \begin{bmatrix}
\mathbf{K}^{1} \\
\mathbf{K}^{2} \\
\vdots \\
\mathbf{K}^{N} \\
\end{bmatrix} = \begin{bmatrix}
\mathbf{\hat{K}}_{TE}^1 & \mathbf{\hat{K}}_{TE}^2 & \dots & \mathbf{\hat{K}}_{TE}^L \\
\mathbf{\hat{K}}_{CL}^1 & \mathbf{\hat{K}}_{CL}^2 & \dots & \mathbf{\hat{K}}_{CL}^L
\end{bmatrix}
\end{align}
where $* \in \{TE, CL\}$ and $N = N_{TE} + N_{CL}$.

We use $\bf{Q}$ and $\bf{K}$ to perform pairwise concatenation to obtain relation resource ${\bf{R}}^{1:L}$~$\in~\mathbb{R}^{(N_{TE} + N_{CL}) \times (N_{TE} + N_{CL}) \times L \times d }$ as in \cite{egtr}:

\begin{align}
    \mathbf{R}^{1:L} =
\begin{bmatrix}
[\mathbf{Q}_1 \; \mathbf{K}_1] & \cdots & [\mathbf{Q}_1 \; \mathbf{K}_N] \\
[\mathbf{Q}_2 \; \mathbf{K}_1] & \cdots & [\mathbf{Q}_2 \; \mathbf{K}_N] \\
\vdots & \ddots & \vdots \\
[\mathbf{Q}_N \; \mathbf{K}_1] & \cdots & [\mathbf{Q}_N \; \mathbf{K}_N]
\end{bmatrix},
\end{align}
where $N = N_{TE} + N_{CL}$ as before.

Similarly, we use the last-layer decoder outputs to form the last-layer relation resource $\mathbf{R}^z$~$\in~\mathbb{R}^{(N_{CL} + N_{TE}) \times (N_{CL} + N_{TE }) \times L \times d}$.

The final relation resource ${\bf{R}}$ is formed by stacking corresponding elements in $\mathbf{R}^{1:l}$ and $\mathbf{R}^{z}$:
\begin{align}
    \mathbf{R} =
\begin{bmatrix}
\begin{bmatrix} \mathbf{R}_{1,1}^{1:L} \\ \mathbf{R}_{1,1}^{z} \end{bmatrix} & \cdots & \begin{bmatrix} \mathbf{R}_{1,N}^{1:L} \\ \mathbf{R}_{1,N}^{z} \end{bmatrix} \\
\begin{bmatrix} \mathbf{R}_{2,1}^{1:L} \\ \mathbf{R}_{2,1}^{z} \end{bmatrix} & \cdots & \begin{bmatrix} \mathbf{R}_{2,N}^{1:L} \\ \mathbf{R}_{2,N}^{z} \end{bmatrix} \\
\vdots & \ddots & \vdots \\
\begin{bmatrix} \mathbf{R}_{N,1}^{1:L} \\ \mathbf{R}_{N,1}^{z} \end{bmatrix} & \cdots & \begin{bmatrix} \mathbf{R}_{N,N}^{1:L} \\ \mathbf{R}_{N,N}^{z} \end{bmatrix}
\end{bmatrix}.
\end{align}

Finally, topology between centerlines are predicted using $\mathbf{R}_{CLCL}$ using gated sums and MLPs as in \cite{egtr}. Similarly, topology between traffic elements and centerlines are predicted using  $\mathbf{R}_{TECL}$. $\mathbf{R}_{CLCL}$ and $\mathbf{R}_{TECL}$ are selected from $\mathbf{R}$:
\begin{align}
\mathbf{R}_{TECL} &= \mathbf{R}_{1:N_{TE},{1:N_{CL}} }\\
\mathbf{R}_{CLCL} &= \mathbf{R}_{(N_{TE}+1):N,{(N_{TE}+1):N}.}
\end{align}
We use the same losses as in \cite{toponet} for detection and topology reasoning.

\begin{table*}[h!]
    \centering
    \begin{tabular}{llccccc}
        \toprule
        & Method & OLS$~\uparrow$ & $\text{DET}_l$~$\uparrow$ & $\text{DET}_t$~$\uparrow$ & $\text{TOP}_{ll}$~$\uparrow$ & $\text{TOP}_{lt}$~$\uparrow$ \\
        \midrule
        \multirow{2}{*}{\textbf{SD-Map-Based}} 
        & SMERF \cite{smerf}  & 43.0  & 31.1  & 48.6  & 16.2  & 27.1 \\
        & Ours (Teacher) & \textbf{44.3}  & \textbf{33.5}  & \textbf{49.4}  & \textbf{17.1}  & \textbf{28.2} \\
        \midrule
        \multirow{6}{*}{\textbf{SD-Map-Free}} 
        & STSU \cite{STSU} & 29.3 & 12.7 & 43.0 & 2.9 & 19.8 \\
        & VectorMapNet \cite{vectormapnet} & 24.9 &11.1 & 41.7 & 2.7 & 9.2  \\
        & MapTR \cite{maptr} & 24.2 & 8.3 & 43.5 & 2.3 & 8.9  \\
        & MapTR (Chamfer Dist.) \cite{maptr} & 31.0 & 17.7 & 43.5 & 5.9 & 15.1  \\
        & TopoNet \cite{toponet}   & 39.8  & 28.6  & 48.6  & 10.9  & 23.8  \\
        & Ours (Student) & \textbf{40.6}  & \textbf{30.2}  & \textbf{48.7}  & \textbf{11.0}  & \textbf{25.2} \\
        \bottomrule
    \end{tabular}
    \caption{Performance comparison on OpenLaneV2 subset-A.}
    \vspace{-15pt}
    \label{tab:results}
\end{table*}


\subsection{Map-to-Mapless Knowledge Distillation}
\label{method_distill}

Our another innovation is distilling the knowledge from map-based models to map-free models (cf. Fig. \ref{architecture}). While knowledge distillation has proven effective in many contexts \cite{zheng2024monoocc}, it has not been explored for transferring knowledge from map-based to map-free models.

To create our teacher network, we apply our proposed method to convert SMERF \cite{smerf} to a one-stage architecture. Compared to TopoNet \cite{toponet}, SMERF is identical except for the cross attention that fuses the extra information from SD maps into the BEV feature. With the extra SD map input, SMERF performs better in terms of detection and topology reasoning. Furthermore, when applied to SMERF, our proposed one-stage method outperforms the original SMERF across all evaluation metrics.

We aim to enable our student network to learn from the superior features of the higher-performing teacher network without relying on SD map inputs. To this end, we enforce similarity between student ($F_{BEV-S}$) and teacher BEV features ($F_{BEV-T}$) using the MSE loss: 
\begin{equation}
L_{\text{BEV}} = \left\| F_{BEV-S} - F_{BEV-T} \right\|_2^2.
\end{equation}

In this way, our student network is optimized jointly using the soft labels generated by the frozen teacher network, in addition to the ground truth.

\section{Experiments}

\subsection{Dataset and Evaluation Metrics} 
\textbf{Datasets.} We train and evaluate our models on the OpenLane-V2 \cite{openlanev2} dataset, the same dataset using by our baselines \cite{toponet, smerf}, including topology annotations that capture relationships between lane centerlines and traffic elements. SD maps are obtained and processed in the same way as in \cite{smerf}. We report results from the subset-A and subset-B. 

\textbf{Metrics.} In line with standard practices, we report the $DET$ score as the Mean Average Precision (mAP) for evaluating instance-level perception performance. Building on Fréchet distances~\cite{eiter1994computing}, the $\text{DET}_{l}$ score is averaged over match thresholds \(\{1.0, 2.0, 3.0\} \). The $\text{DET}_{t}$ score, using Intersection over Union (IoU) as the similarity measure, is averaged across thirteen attributes of traffic elements. For topology evaluation, we employ the official $\text{TOP}_{ll}$ metric to assess mAP for lane centerline topology, the $\text{TOP}_{lt}$ metric for topology between lane centerlines and traffic elements, and  the overall metric OpenLane-V2 Score (OLS) from~\cite{openlanev2}.

\subsection{Implementation Details} 
We train our models using four A800 or four V100 GPUs, with a batch size of 1 per card. We employ The AdamW optimizer and the initial learning rate is \( 1 \times 10^{-4} \). The training is carried out for a total of 24 epochs. The number of transformer decoder layers and queries are the same as in TopoNet~\cite{toponet} and SMERF~\cite{smerf}. 

Inference speeds are measured on a machine with four V100 GPUs, using only one GPU while the other three remain idle. Feature extraction time is excluded from speed comparisons unless otherwise specified.


\begin{table}[h!]
    \centering
    \resizebox{\columnwidth}{!}{ 
    \begin{tabular}{llcc}
        \toprule
        & Method & Inference Speed~$\uparrow$ & \#~parameters~$\downarrow$ \\
        \midrule
        \multirow{2}{*}{\textbf{SD-Map-Based}}  
        & SMERF \cite{smerf}  & +0\% & 65.8M \\
        & Ours (Teacher) & \textbf{+17\%} & \textbf{52.7M} \\
        \midrule
        \multirow{2}{*}{\textbf{SD-Map-Free}}  
        & TopoNet \cite{toponet}  & +0\% & 62.6M \\
        & Ours (Student) & \textbf{+17\%} & \textbf{49.4M} \\
        \bottomrule
    \end{tabular}
    }
    \caption{Inference speed gain and parameters comparison.}
    \vspace{-0.5cm}
    \label{tab:results_speed}
\end{table}

\subsection{Quantitative Results} 
We compare the performance of our proposed one-stage architecture with two-stage state-of-the-art methods in \autoref{tab:results}. Results of STST~\cite{STSU}, VectorMapNet~\cite{vectormapnet}, MapTR~\cite{maptr}, and MapTR (Chamfer Dist.)~\cite{maptr} are taken from TopoNet \cite{toponet}. Their inference speed and parameter counts are not available because \cite{toponet} modified those methods to get these results but did not release relevant code. When SD maps are available, our one-stage teacher network outperforms SMERF~\cite{smerf}. In the absence of SD maps, our distilled one-stage student network surpasses TopoNet. In addition to higher accuracy, Table~\ref{tab:results_speed} shows that our one-stage method is up to 17\% faster for inference and have less model parameters.

Specifically, as shown in the upper half of \autoref{tab:results}, compared to the SD-map baseline \cite{smerf}, our teacher network improves the overall score, traffic elements detection accuracy, and topology reasoning accuracy between lane centerlines and traffic elements by 1 point, while increasing lane detection accuracy and topology reasoning accuracy among lanes by more than 2 points. Without SD map inputs, our distilled student network achieves superior overall performance, enhancing traffic element detection accuracy and topology reasoning accuracy among lanes by more than 1 point, when compared to the SD-map-free baseline \cite{toponet}.

We additionally report our results on subset-B of the OpenLane-V2 dataset in \autoref{tab:subset_b}. Even without knowledge distillation, our method outperforms the state-of-the-art method TopoNet, achieving a  higher score in overall metric.

\begin{table}[h!]
    \centering
    \resizebox{\columnwidth}{!}{
    \begin{tabular}{lccccc}
        \toprule
        Method & OLS & $\text{DET}_l$~$\uparrow$ & $\text{DET}_t$~$\uparrow$ & $\text{TOP}_{ll}$~$\uparrow$ & $\text{TOP}_{lt}$~$\uparrow$ \\
        \midrule
        TopoNet \cite{toponet}   & 36.0  & 24.4  & 52.6  & 6.7  & \textbf{16.7}  \\
        Ours (No Distillation) & \textbf{37.0}  & \textbf{25.4}  & \textbf{55.5}  & \textbf{6.9}  & 16.5  \\
        \bottomrule
    \end{tabular}
    }
    \caption{Performances on OpenLaneV2 subset-B.}
    \vspace{-0.5cm}
    \label{tab:subset_b}
\end{table}

\begin{figure*}[h]
  \centering
   \includegraphics[width=0.75\linewidth]{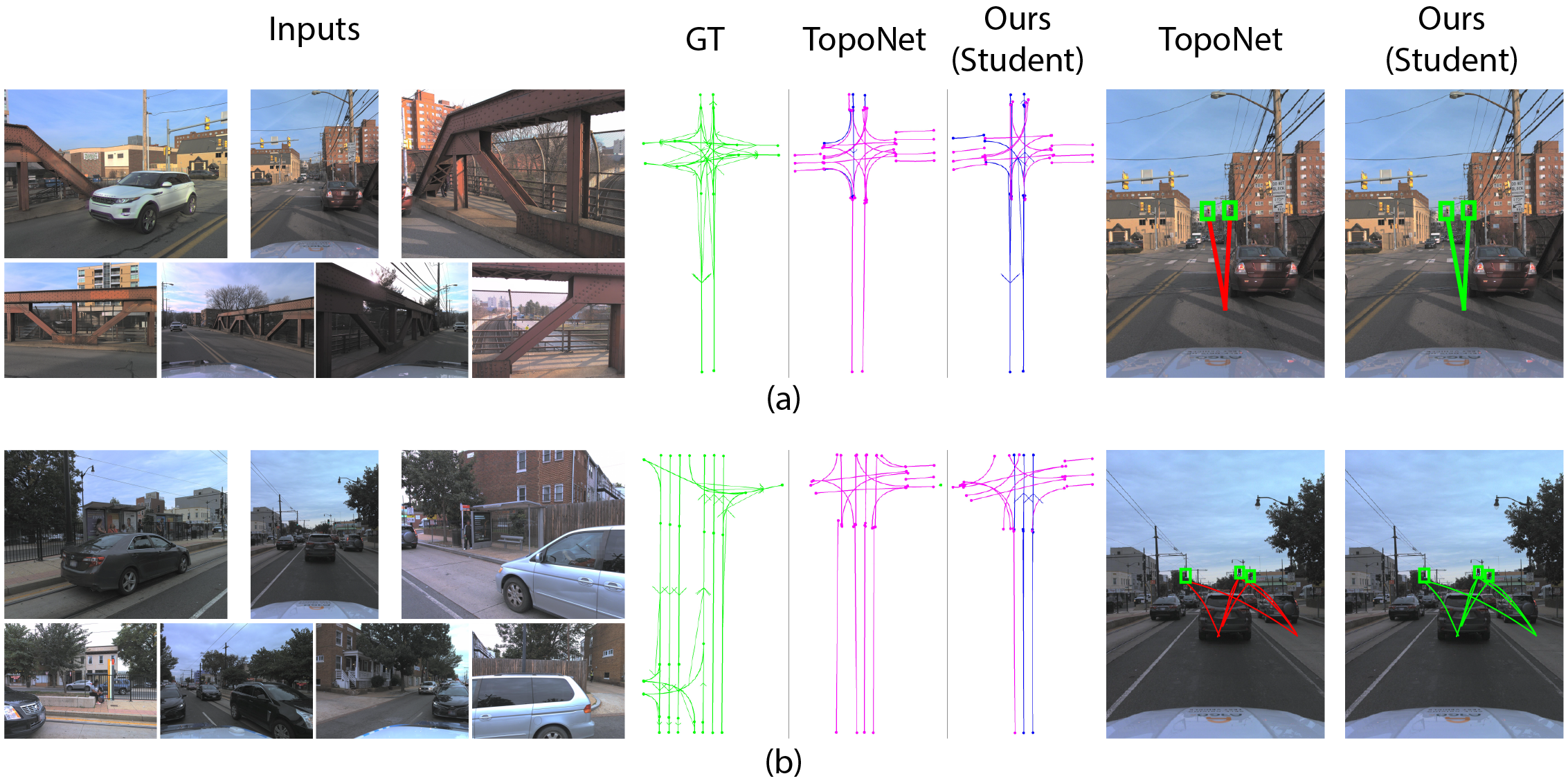}
   \caption{
   \textbf{Qualitative Comparisons} between TopoNet~\cite{toponet} and our student network. 
   \textbf{Left (Multi-View Inputs)}: Visualization of corresponding multi-view inputs.
   \textbf{Middle (CL and CLCL Predictions)}: Purple indicates \textcolor[RGB]{226,49,243}{false positives}, while blue denotes \textcolor[RGB]{51,49,220}{true positives}.  
   \textbf{Right (TE and TECL Predictions)}: Green represents \textcolor{green}{true positives}, whereas red signifies \textcolor{red}{false negatives}.
   }
   \vspace{-0.2cm}
   \label{qualitative_student}
\end{figure*}

\begin{figure*}[h]
  \centering
   \includegraphics[width=0.75\linewidth]{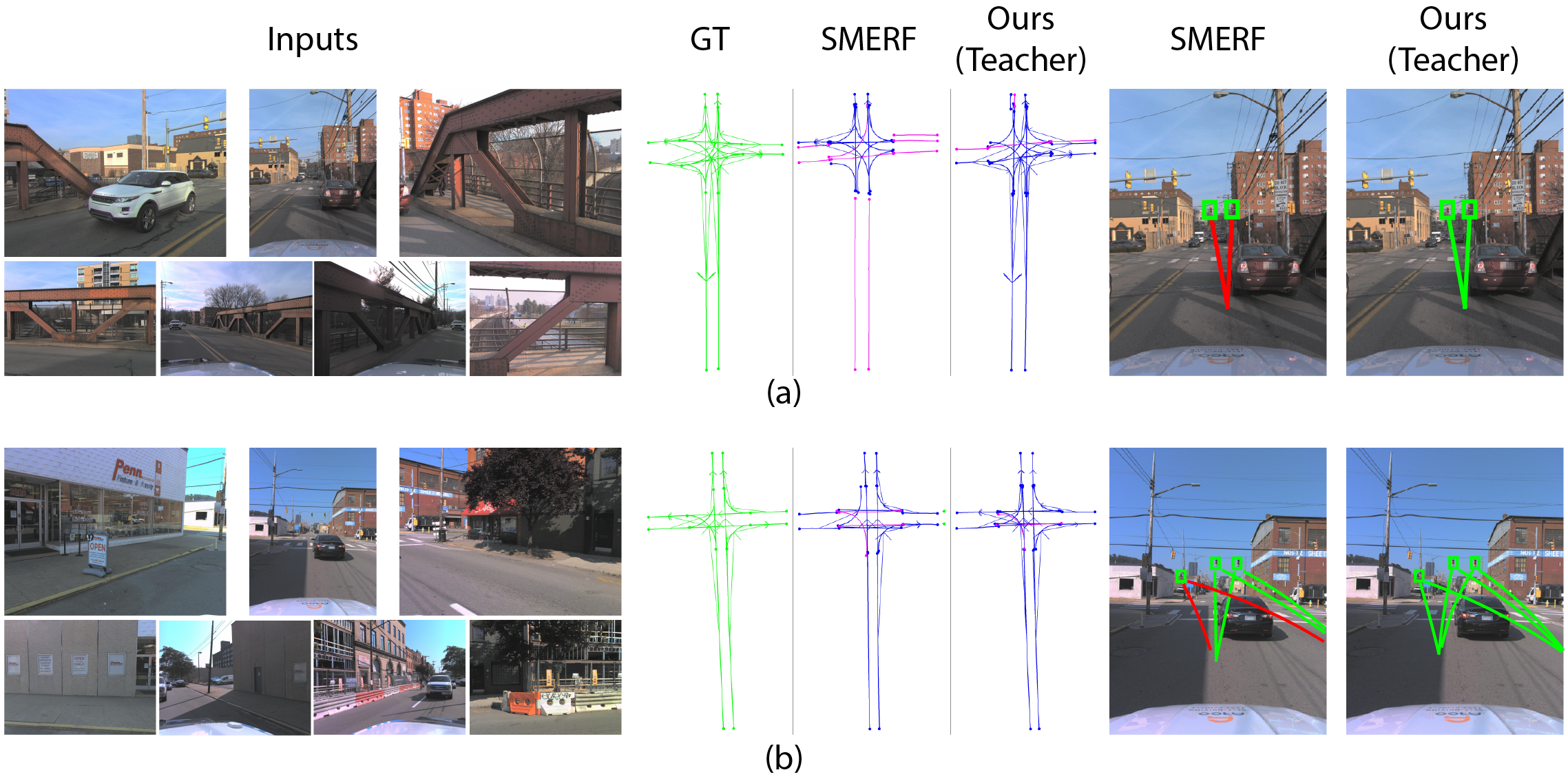}
   \caption{
   \textbf{Qualitative Comparisons} between TopoNet~\cite{toponet} and our student network. 
   \textbf{Left (Multi-View Inputs)}: Visualization of corresponding multi-view inputs.
   \textbf{Middle (CL and CLCL Predictions)}: Purple indicates \textcolor[RGB]{226,49,243}{false positives}, while blue denotes \textcolor[RGB]{51,49,220}{true positives}.  
   \textbf{Right (TE and TECL Predictions)}: Green represents \textcolor{green}{true positives}, whereas red signifies \textcolor{red}{false negatives}.
   }
   \vspace{-10pt}
   \label{qualitative_teacher}
\end{figure*}


\begin{figure}[h]
  \centering
   \includegraphics[width=0.95\linewidth]{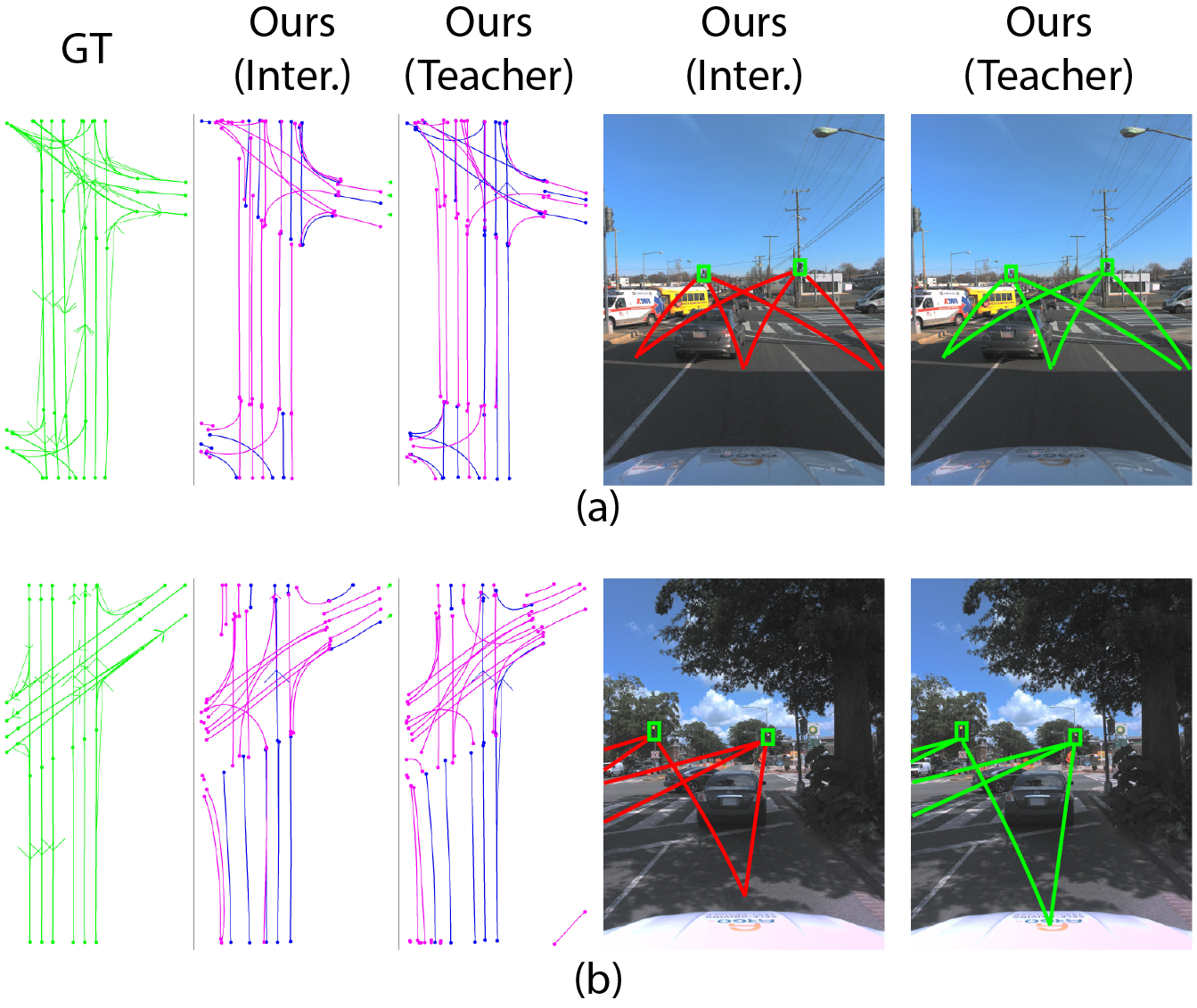}
   \caption{\textbf{Qualitative Comparisons} between our network with interactions and our teacher network.  
\textbf{Left (CL and CLCL Predictions)}: Purple indicates \textcolor[RGB]{226,49,243}{false positives}, while blue denotes \textcolor[RGB]{51,49,220}{true positives}.  
\textbf{Right (TE and TECL Predictions)}: Green represents \textcolor{green}{true positives}, whereas red signifies \textcolor{red}{false negatives}.  
}
\vspace{-0.6cm}
   \label{qualitative_conv}
\end{figure}

\subsection{Qualitative Comparisons} 





\subsubsection{Without SD Map Inputs}
When SD map inputs are not available, our student network performs better than TopoNet. In Fig.~\ref{qualitative_student}, we present qualitative results from left to right: multi-view image inputs, BEV visualizations of LC-LC topology predictions, and TE detection predictions and LC-TE topology predictions. When predicting LC-LC topology, our student network outperforms TopoNet by generating more true positives, particularly around the lanes the ego vehicle occupies.

Our student network also excels in reasoning about LC-TE topology. In Fig.~\ref{qualitative_student} (a), right columns, both TopoNet and our student network correctly detect the traffic lights highlighted in green boxes. However, as indicated by the red lines, TopoNet fails to infer the topological relationships between these traffic lights and the lane of the ego vehicle. In contrast, our student network successfully captures these relationships.

A similar pattern emerges in Fig.~\ref{qualitative_student} (b), right columns, where both networks correctly detect the three traffic lights marked in green. However, TopoNet fails to recognize all seven LC-TE relationships (marked by red curves), whereas our student network successfully predicts all of them (marked by green curves).

These results demonstrate that our proposed one-stage architecture, enhanced with knowledge distilled from an SD-map-based teacher network, is more effective at detecting lanes and reasoning about LC-LC and LC-TE relationships.





\subsubsection{With SD Map Inputs}
With SD map inputs, our teacher network outperforms SMERF, which also utilizes SD maps. In Fig.~\ref{qualitative_teacher}, we present qualitative comparisons between SMERF and our teacher network.

Thanks to the additional information provided by SD maps, both networks, as shown in the middle columns of Fig.~\ref{qualitative_teacher}, perform better at predicting LC-LC topology compared to models without SD map inputs (Fig.~\ref{qualitative_student}). However, our teacher network surpasses SMERF by correctly predicting a greater number of LC-LC relationships.

Our teacher network also demonstrates superior performance in LC-TE topology reasoning. As shown in the right columns of Fig.~\ref{qualitative_teacher}, it correctly detects all LC-TE relationships (represented by green lines), whereas SMERF misses two (highlighted in red) in both (a) and (b).

These results, along with higher scores in Table~\ref{tab:results}, confirm that our proposed one-stage architecture further enhances the topological reasoning capabilities of SD-map-based models.

\subsection{Ablation}
\subsubsection{Effects of Distillation}

We compare the performances of our models with and without distillation when SD maps are not available in \autoref{tab:distillation}.

Without distillation, our one-stage model achieves the same overall accuracy (OLS) as TopoNet \cite{toponet} while being 17\% faster. After distilling knowledge from the teacher network trained with SD maps, our student model retains the 17\% speed advantage while surpassing TopoNet in accuracy.

The improved accuracy after distillation demonstrates the effectiveness of transferring knowledge from SD-map-based models to SD-map-free models, enhancing performance even in the absence of one input modality.

\begin{table}[h!]
    \centering
    \resizebox{\columnwidth}{!}{
    \begin{tabular}{lccccc}
        \toprule
        Method & OLS & $\text{DET}_l$~$\uparrow$ & $\text{DET}_t$~$\uparrow$ & $\text{TOP}_{ll}$~$\uparrow$ & $\text{TOP}_{lt}$~$\uparrow$\\
        \midrule
        Ours (No distillation)      & 39.9  & 29.6  & 47.8  & 10.5  & 24.7 \\
        Ours (Student) & \textbf{40.6}  & \textbf{30.2}  & \textbf{48.7}  & \textbf{11.0}  & \textbf{25.2}  \\
        \bottomrule
    \end{tabular}
    }
    \caption{Effects of Distillation.}
    \label{tab:distillation}
    \vspace{-15pt}
\end{table}

\subsubsection{Effects of Extra Feature Interactions}




To investigate the impact of feature interactions on model accuracy, we compared the performances of our models with and without enabling feature interactions.

In our experiments, we allowed extra feature interactions by utilizing the complete matrix $\mathbf{R}$ instead of just task-relevant $\mathbf{R}_{CLCL}$. This resulted in a feature set of size $(N_{TE} + N_{CL}) \times (N_{TE} + N_{CL})$, as opposed to restricting our focus to the $N_{CL} \times N_{CL}$ task-relevant features typically used for reasoning topology among lanes. To facilitate these feature interactions, we applied a 2D convolution with a kernel size of 3 and a stride of 1, using the ``same'' padding technique. This was followed by a 2D adaptive average pooling operation, which reduced the output to the desired shape of $N_{CL} \times N_{CL}$ from the larger feature set.

Contrary to expectations, as illustrated in \autoref{tab:feature_interaction}, enabling extra feature interactions led to a notable degradation in accuracy across all evaluation metrics. This suggests that, at least in our case, the inclusion of additional feature interactions may not be beneficial for model performance in the context of topology reasoning.

\begin{table}[h!]
    \centering
    \resizebox{\columnwidth}{!}{
    \begin{tabular}{lccccc}
        \toprule
        Method & OLS & $\text{DET}_l$~$\uparrow$ & $\text{DET}_t$~$\uparrow$ & $\text{TOP}_{ll}$~$\uparrow$ & $\text{TOP}_{lt}$~$\uparrow$\\
        \midrule
        Ours (Teacher) & \textbf{44.3}  & \textbf{33.5}  & \textbf{49.4}  & \textbf{17.1}  & \textbf{28.2}\\
         Ours (Interactions) & 42.9  & 33.0  & 47.1  & 16.6  & 25.8\\
        \bottomrule
    \end{tabular}
    }
    \caption{Effects of Feature Interactions.}
    \vspace{-0.4cm}
    \label{tab:feature_interaction}
\end{table}

We also present qualitative comparisons between our teacher network and our network with additional feature interactions in Fig.~\ref{qualitative_conv}, further confirming that extra feature interactions negatively impact the model performance.

\section{CONCLUSIONS}


In this paper, we address the limitations of current two-stage frameworks in road topology understanding for autonomous driving. Our novel single-stage road topology reasoning architecture integrates instance detection and relation prediction across both Perspective View and Bird’s-Eye View, improving performance and efficiency. Additionally, our Map-to-Mapless Knowledge Distillation method transfers knowledge from a high-performance, map-based teacher model to a lightweight, camera-only student model, enhancing road topology reasoning accuracy without compromising efficiency. Experimental results on real-world data show that our approach surpasses state-of-the-art methods in both accuracy and inference speed.





\bibliographystyle{IEEEtran}
\bibliography{main}

\end{document}